\documentclass[runningheads]{llncs}

\usepackage{eccv}

\usepackage{eccvabbrv}

\usepackage{wrapfig}
\usepackage{graphicx}
\usepackage{booktabs}
\usepackage{multirow}
\usepackage[accsupp]{axessibility}  % Improves PDF readability for those with disabilities.

\usepackage{hyperref}

\usepackage{orcidlink}

\newcommand{\ourmodel}{HairWeaver\xspace}
\newcommand{\pose}{Motion-Context-LoRA\xspace}
\newcommand{\domain}{Style-Alignment-LoRA\xspace}

\usepackage{xcolor,soul}
\sethlcolor{yellow}

\begin{document}

% ---------------------------------------------------------------
% TODO REVIEW: Replace with your title
\title{\ourmodel\texorpdfstring{\includegraphics[height=1.2em]{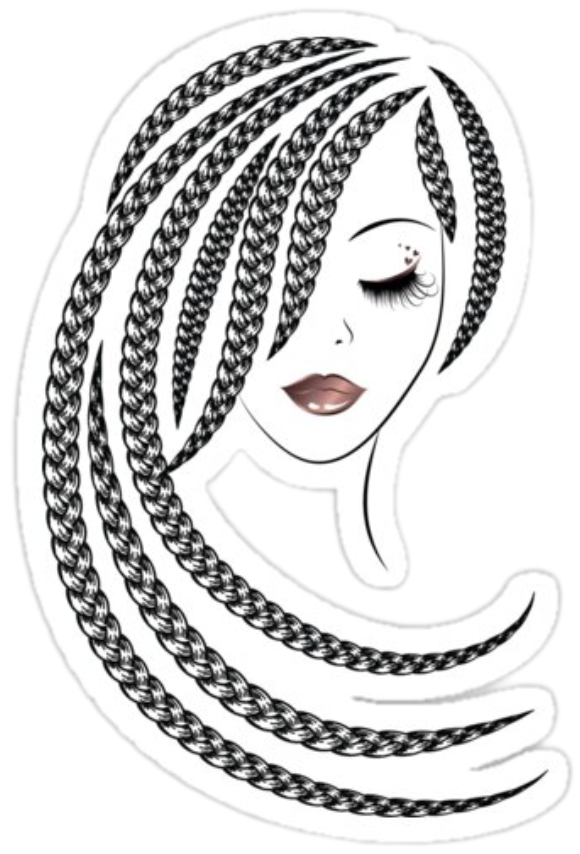}}{[hair emoji]}: Photorealistic Hair Motion Synthesis with Sim-to-Real Physics Transfer Guided Video Diffusion}

% TODO REVIEW: If the paper title is too long for the running head, you can set
% an abbreviated paper title here. If not, comment out.
\titlerunning{\ourmodel}

% TODO FINAL: Replace with your author list. 
% Include the authors' OCRID for the camera-ready version, if at all possible.
\author{Di Chang\inst{1,2}\orcidlink{0009-0002-0281-8896} \and
Ji Hou\inst{1}\orcidlink{0000-0002-5244-8953} \and
Aljaz Bozic\inst{1}\orcidlink{0009-0002-2985-6921} \and
Assaf Neuberger\inst{1}\and \\
Felix Juefei-Xu\inst{1}\orcidlink{0000-0002-0857-8611} \and
Olivier Maury\inst{1}\orcidlink{0009-0004-3985-1934} \and
Gene Wei-Chin Lin\inst{1} \and
Tuur Stuyck\inst{1}\orcidlink{0000-0003-1892-2137} \and
Doug Roble\inst{1}\orcidlink{0009-0004-3415-4283} \and
Mohammad Soleymani\inst{2}\orcidlink{0000-0002-5873-1434} \and
Stephane Grabli\inst{1}\orcidlink{0009-0006-5616-1753}}

% TODO FINAL: Replace with an abbreviated list of authors.
\authorrunning{Di Chang et al.}
% First names are abbreviated in the running head.
% If there are more than two authors, 'et al.' is used.

% TODO FINAL: Replace with your institution list.
\institute{Meta \and University of Southern California\\
\url{https://boese0601.github.io/hairweaver/} \\
\email{dichang@usc.edu}}

\maketitle

\begin{abstract} 
\label{abstract}
We present \ourmodel, a diffusion-based pipeline that animates a single human image with realistic and expressive hair dynamics. While existing methods successfully control body pose, they lack specific control over hair, and as a result, fail to capture the intricate hair motions, resulting in stiff and unrealistic animations. \ourmodel~overcomes this limitation using two specialized modules: a \textbf{\pose} to integrate motion conditions and a \textbf{\domain} to preserve the subject's photoreal appearance across different data domains.
These lightweight components are designed to guide a video diffusion backbone while maintaining its core generative capabilities. By training on a specialized dataset of dynamic human motion generated from a CG simulator, \ourmodel~affords fine control over hair motion and ultimately learns to produce highly realistic hair that responds naturally to movement. Comprehensive evaluations demonstrate that our approach sets a new state of the art, producing lifelike human hair animations with dynamic details.
\end{abstract}

\vspace{-5pt}
\section{Introduction}
\label{intro}

This work addresses the challenging problem of animating a single source image driven by a motion sequence. This technology holds immense promise for applications in virtual reality, next-generation gaming, and the film industry~\cite{metaverse2023,unreal_engine_5_2024}. While recent progress in generative models has enabled plausible human animation, a significant hurdle remains in synthesizing the intricate, non-rigid dynamics of secondary elements, most notably hair. The complex physics and fine-grained visual nature of hair motion are often overlooked by these models, leading to generated humans with fluid-like and unnatural hair motions, shattering the illusion of realism. Building on prior research in human animation~\cite{hu2024animate,xu2024magicanimate,chang2023magicpose,wang2024vividpose}, we aim to explicitly model and generate expressive and physically plausible hair dynamics in concert with accurate body pose transfer.

Recent breakthroughs in human image animation~\cite{chang2023magicpose,zhang2024mimicmotion, hu2024animate,wang2025unianimate,wang2025dreamactor,chang2025x,zhao2025x,song2025x,zhang2025x,wang2025dreamactor,luo2025dreamactor,chen2025x,wang2024vividpose} are largely driven by controlled image-to-video diffusion models~\cite{zhang2023adding,guo2023animatediff}. These methods typically disentangle appearance and motion by conditioning a pretrained video diffusion model on a reference image for appearance and a sequence of poses (e.g., Pose skeletons~\cite{chang2023magicpose,zhang2024mimicmotion, hu2024animate,wang2025unianimate,zhang2025x,luo2025dreamactor,wang2025dreamactor} or DensePose~\cite{xu2024magicanimate}) for motion control. Appearance is often preserved using attention-based mechanisms that inject features from the reference image into the generation process~\cite{cao2023masactrl,blattmann2023align}. Despite achieving impressive control over body posture and identity preservation, these models fundamentally struggle with secondary dynamics. Their network architectures and training paradigms are optimized for overall postural correctness, treating hair as a static texture mapped to the head.

Alternatively, 3D-based models~\cite{lin2025neuralocks, stuyck2025quaffure, li2025self, zhang2025hairformer, wang2023neuwigs, wang2022hvh}, offer a potential solution to generate diverse hair motions with controllability. However, these methods often lack the photorealism and fine details, producing results that appear synthetic and cannot be directly used to train a diffusion model as previous human video animation methods. We argue that these animation methods falter for a common reason: a critical lack of high-fidelity paired data that contain photorealistic appearance, and diverse, controllable, physically-plausible hair geometry conditions.

To solve this, we propose \ourmodel, a framework that achieves both fine-grained control and photorealism by addressing the data problem head-on. We first generate a small-scale synthetic dataset using a physics-based method~\cite{kugelstadt2016position}, providing rich, physically-grounded hair motion and body pose conditions. We then propose a two-stage simulation-to-real physics transfer training strategy to transfer this motion control to a powerful, pre-trained photorealistic video diffusion transformer (DiT)~\cite{peebles2023scalable} \textit{\textbf{without}} transferring the CG domain's non-photorealistic style. This strategy involves two lightweight components: a \textbf{\pose} module to inject the new motion conditions, and a temporary \textbf{\domain} module to overcome the domain gap between CG simulated and photorealistic videos. The \domain is first trained to adapt the DiT to the CG domain. Then, it is frozen, and the \pose is trained to learn the mapping from motion conditions to video. Crucially, the \domain is discarded entirely during inference. This strategy allows \ourmodel to leverage the precise motion physics from the CG data while retaining the full photorealistic power of the original foundation model.

As a result, \ourmodel achieves a unique combination of capabilities: the fine-grained, controllable hair physics, and the high-fidelity photorealism, while being trained on only a few samples (around 1k videos). We conduct comprehensive evaluations against state-of-the-art baselines on challenging video benchmarks. \ourmodel consistently outperforms existing methods in both quantitative metrics and qualitative user studies, particularly in the realism of motion and the preservation of fine details. In summary, our main contributions are:
\begin{itemize}
    \item A novel diffusion-based pipeline, \ourmodel, specifically designed to synthesize expressive and dynamic hair motion for realistic simulation-to-real physics transfer human video animation, with a synthetic guiding signal from a simulator as training data.
    \item An efficient and effective \textbf{\pose} that injects hair motion control as additional attention context, preserving the generative power of the video diffusion backbone.
    \item A two-stage training strategy that uses a temporary \textbf{\domain} to learn motion control from synthetic data, which is then discarded at inference to ensure photorealistic generation.
\end{itemize}
\begin{figure*}[]\vspace{-15pt}
\centering
 \includegraphics[width=\linewidth]{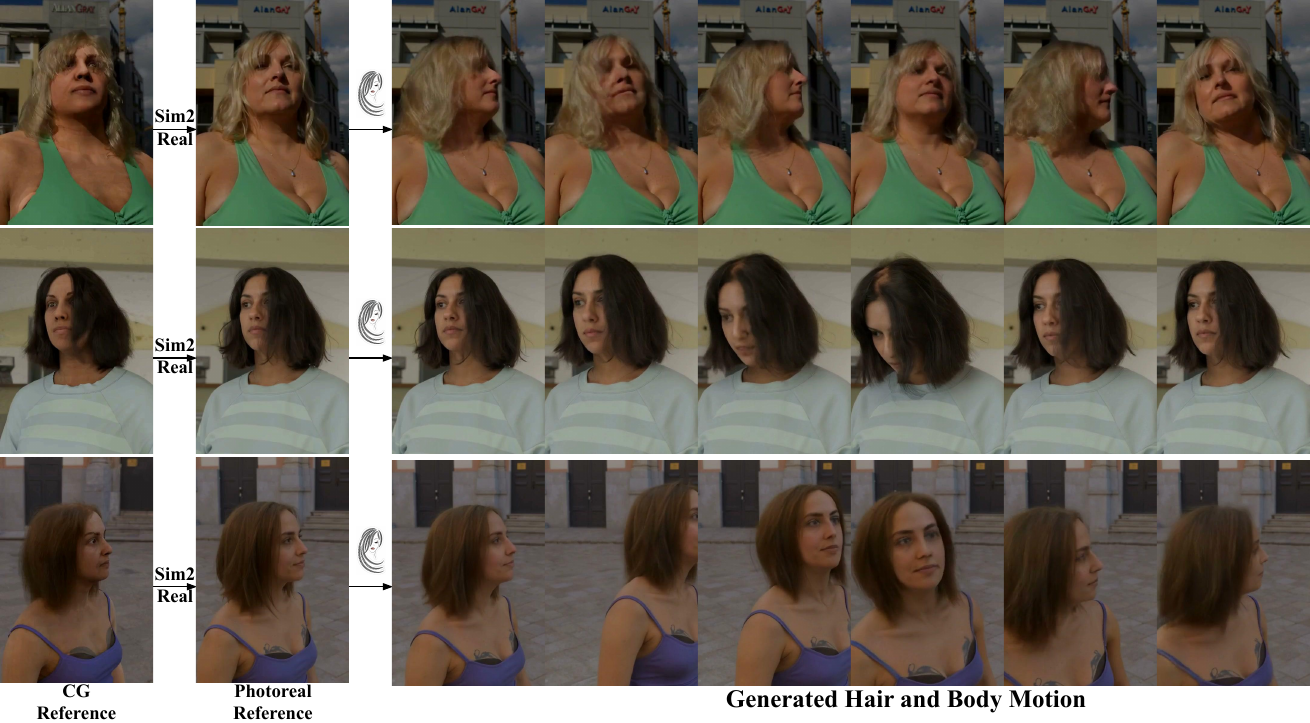}
  \vspace{-5pt}
   \captionof{figure}{Sim-to-real hair animation pipeline. From left to right: a CG-rendered reference, its photorealistic counterpart obtained via image stylization, and frames from the resulting animation generated by HairWeaver with physics-based hair dynamics.
   %\captionof{figure}{Sim-to-real hair animation pipeline. From left to right: a CG-rendered reference, its photorealistic counterpart obtained via image stylization, and frames from the resulting animation generated by \ourmodel with physics-based hair dynamics. 
   }
   \vspace{-15pt}
    \label{fig:teaser}
\end{figure*}

\begin{figure*}[t!]\vspace{-5pt}
\centering
 \includegraphics[width=\linewidth]{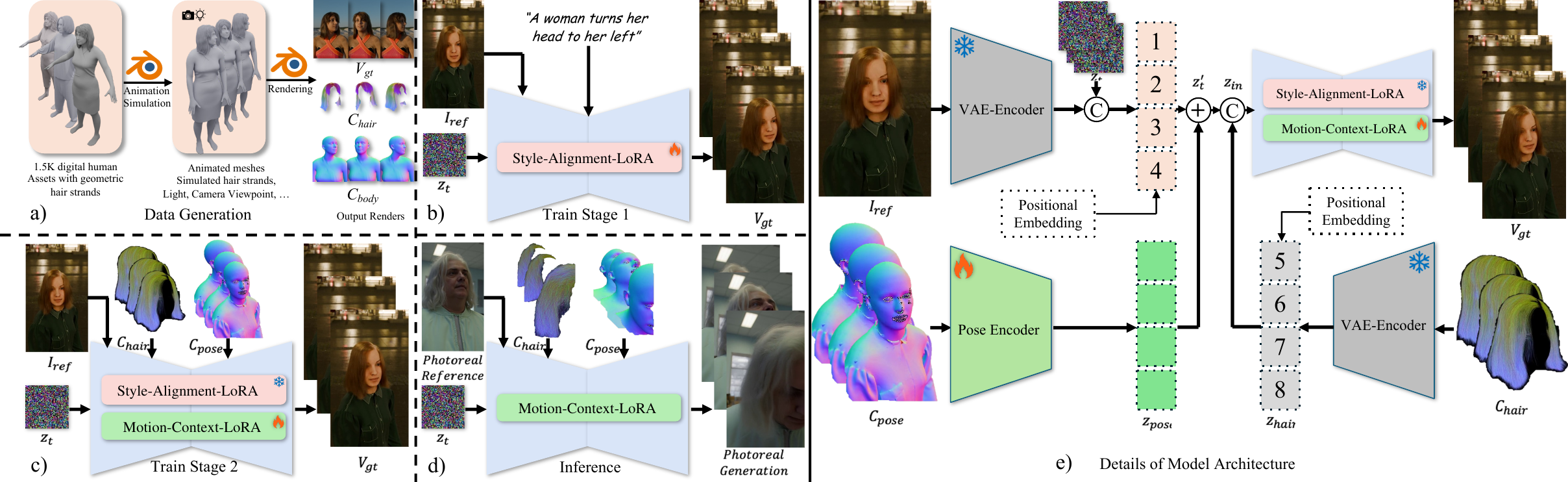}
  \vspace{-13pt}
   \caption{ Overview of \ourmodel pipeline. \textbf{a)} We use CG simulation to generate data including human videos with motions $\mathbf{V}_{gt}$, static reference image $\mathbf{I}_{ref}$ (a frame from $\mathbf{V}_{gt}$), pose condition $\mathbf{C}_{pose}$, and hair condition $\mathbf{C}_{hair}$. \textbf{b)} During training stage 1, we leverage the a diffusion transformer~\cite{peebles2023scalable} (DiT)  as the backbone model and pre-train the \domain. This training process is conducted in Image-to-Video manner with $\mathbf{I}_{ref}$ and text prompt for $\mathbf{V}_{gt}$. \textbf{c)} During training stage 2, we freeze the \domain and finetune the \pose with $\mathbf{C}_{pose}$, and hair condition $\mathbf{C}_{hair}$ as additional guidance. 
   \textbf{d)} During inference, the \domain is discarded and the trained model generates photorealistic human videos with hair and body motions with photorealistic reference and CG conditions $\mathbf{C}_{pose}, \mathbf{C}_{hair}$ as input.
   \textbf{e)} Details of the model architecture presented in (c). The Pose Encoder integrates the body motions as a trainable residual to the noisy latent. The hair motions are encoded as additional attention context to the DiT blocks by a frozen VAE-Encoder. The only trainable modules are the Pose Encoder and the \pose. }
    \vspace{-10pt}
    \label{fig:pipeline}
\end{figure*}

\section{Related Work}
\label{sec:related}

\subsection{Diffusion Models for Human Video Animation}
The animation of human subjects from a single image has been revolutionized by latent diffusion models~\cite{rombach2022high}. A prevalent paradigm involves conditioning a model on both appearance and motion. To preserve the subject's identity, methods often employ a ReferenceNet architecture or cross-attention mechanisms to inject appearance features from a source image into the generation process~\cite{hu2024animate,chang2023magicpose,zhu2024champ,cao2023masactrl,ye2023ip}. For motion control, pose information derived from a driving video is typically supplied through a spatial guidance module like ControlNet~\cite{zhang2023adding,xu2024magicanimate,chang2023magicpose}.

Before video models became ubiquitous, many methods followed a two-stage training process: first training a static image generator and then adding a temporal module for video consistency~\cite{wang2023disco,chang2023magicpose}. However, the advent of powerful, large-scale video foundation models has enabled a more streamlined approach. Recent state-of-the-art works, such as MimicMotion~\cite{zhang2024mimicmotion} and UniAnimate-DiT~\cite{wang2025unianimate}, directly fine-tune Video Diffusion models like Stable Video Diffusion (SVD)~\cite{blattmann2023stable} or Wan2.1~\cite{wan2025}. By leveraging the strong temporal priors of these pretrained models, they achieve high-fidelity motion transfer and identity preservation. Despite their success in replicating overall body movements, these frameworks are not explicitly designed to capture fine-grained secondary dynamics, a limitation that becomes particularly apparent in complex materials like hair.

\subsection{Hair Synthesis}
Synthesizing highly convincing, photorealistic human hair --- particularly in dynamic scenarios --- remains a significant challenge.
Over the past several decades, the Visual Effects (VFX), Animation, and Video Game industries have refined the craft of hair synthesis using Computer Graphics (CG), both in offline rendering and real-time settings.
CG-based hair offers numerous advantages. By representing hair as hundreds of thousands of individual geometric strands, artists gain a high degree of control and flexibility in the creative process.
Skilled artists achieve compelling hair renders by leveraging advanced grooming and simulation tools~\cite{SideFX:Houdini}, sophisticated hair shading models~\cite{chiang2015practical}, and physically-based path tracing algorithms~\cite{pharr2023physically}.
Despite the relative ease of producing high-quality CG hair, achieving true photorealism remains elusive. Only top-tier studios can generate truly realistic video sequences, and doing so incurs substantial costs. One of our goals is to harness the controllability of CG hair combined with video diffusion models, to achieve photorealistic results on par with, or surpassing, high-end productions, while significantly reducing costs.

As previously discussed, while video diffusion models excel at producing realistic results and provide some degree of control over body pose, they generally lack explicit mechanisms for directing hair synthesis. This limitation poses a significant challenge when creating movies that require detailed and controllable hair dynamics. One notable exception is the recent 
ControlHair~\cite{lin2025controlhair} paper, which finetunes a video diffusion backbone~\cite{wan2025} to offer directable  hair motion. 
ControlHair utilizes HairStep's hair direction neural extractor ~\cite{zheng2023hairstep}
to generate its driving signals. However, unlike CG data --- which offers perfect alignment between conditioning and target signals, albeit with a domain gap --- these extracted maps are only approximate. This limited accuracy ultimately constrains the quality and expressiveness of the resulting model. We choose to rely more thoroughly on CG data and describe our solution to handling 
its domain gap in section~\ref{sec:domain_lora}.

% As mentioned before, while video diffusion models excel at realism and afford some control over body pose, no explicit control typically exists to direct hair synthesis. We argue that this lack of control 
% can represent a challenge for making movies featuring prominent 
% hair dynamics. One notable exception is the recent 
% ControlHair~\cite{lin2025controlhair} paper, which finetunes a video diffusion backbone~\cite{wan2025} to offer directable  hair dynamics. However, their use of a the accuracy it still cannot generate wild hair motions with good accuracy, due to the limitation of the hair motion detector~\cite{zheng2023hairstep}.

Recently, 3D Gaussian Splatting (3DGS)~\cite{kerbl2023_3dgs} has emerged as an innovative 3D representation, offering a promising shortcut to achieving photorealism.
However, this approach trades off versatility for visual fidelity: the photorealistic appearance is typically derived from static multi-view captures and baked into the representation. As a result, animating or relighting objects represented with 3DGS remains a challenge. Notably, recent work such as Gaussian Haircut~\cite{zakharov2024human} has sought to bridge this gap by integrating 3DGS with traditional strand-based hair models, aiming to restore the flexibility inherent in classic approaches. Nevertheless, animating hair within the 3DGS framework continues to be an open research problem.

\begin{figure*}[t!]\vspace{-5pt}
\centering
 \includegraphics[width=0.99\linewidth]{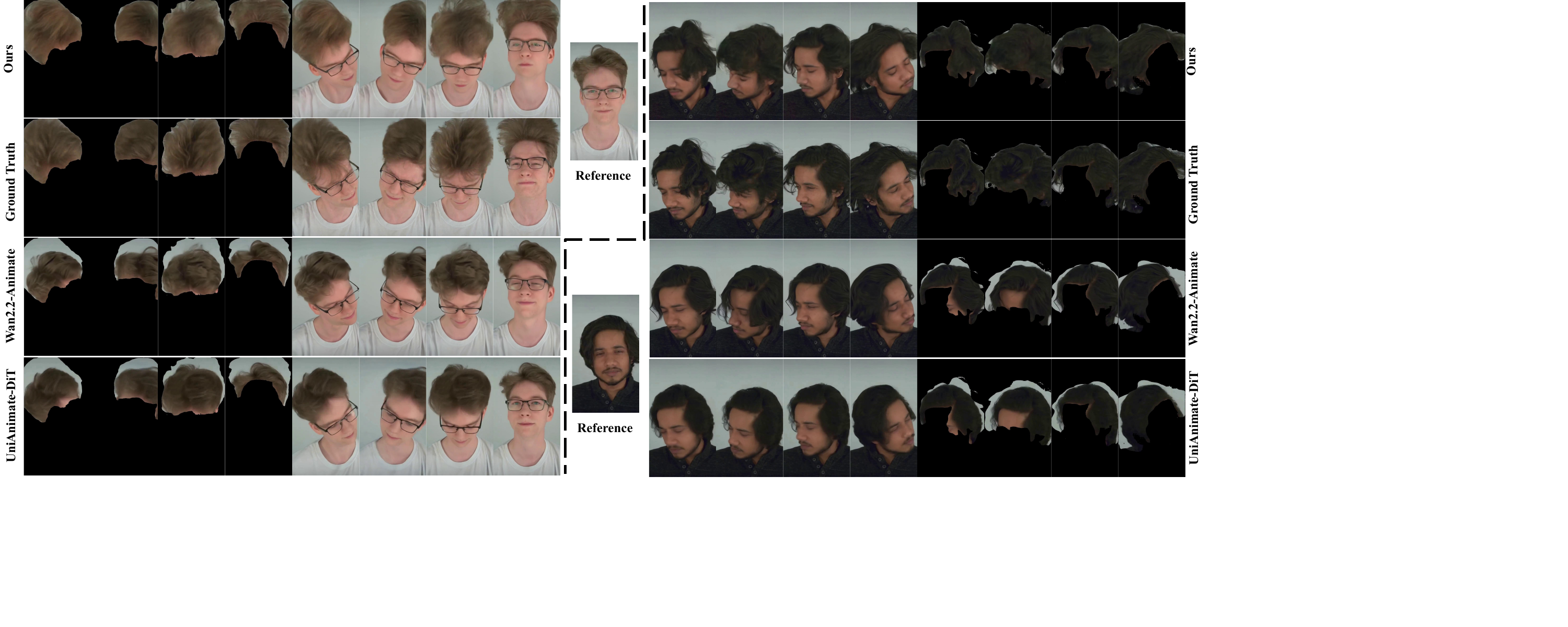}
  \vspace{-5pt}
   \caption{{ Visualization of comparison between \ourmodel and the previous state-of-the-art baselines~\cite{wang2025unianimate,wan2025}. Our model generates more realistic and diverse hair motions.
   }
   }
   \vspace{-5pt}
    \label{fig:ns_comp}
\end{figure*}

\section{Method}\label{sec:method}
Our method involves two main stages. First, we generate a synthetic dataset by rendering a set of animated digital human assets, where the hair is represented as geometric strands and animated using physics-based simulation, providing rich and physically-grounded hair motion signals paired with body pose, as detailed in Section~\ref{sec:data_gen}. Second, we propose a novel two-step training strategy to adapt a large-scale Video DiT~\cite{peebles2023scalable} for this task without inheriting the synthetic data's non-photorealistic artifacts. This strategy involves \pose, a lightweight module for injecting our dual motion conditions (described in Section~\ref{sec:pose_lora}), and \domain, a temporary LoRA that bridges the domain gap during training (described in Section~\ref{sec:domain_lora}). Crucially, \domain is discarded at inference, allowing \pose to drive the original photorealistic backbone. This results in a model with the fine-grained controllability of our CG data and the high-fidelity realism of the pre-trained DiT. The pipeline of \ourmodel is depicted in Fig.~\ref{fig:pipeline}.

\subsection{Preliminaries}\label{sec:pre}

The video diffusion transformer (DiT) \cite{peebles2023scalable} represents a significant advancement in video generation, utilizing fully transformer-based models in diffusion frameworks. 

The core principle of DiT involves a forward diffusion process that adds Gaussian noise to the data, and a reverse denoising process that learns to reconstruct the original video from the noisy input. Given a video sequence $\mathbf{x}_0 \in \mathbb{R}^{T \times H \times W \times C}$, where $T$ is the number of frames, $H$ and $W$ are spatial dimensions, and $C$ is the number of channels, the forward process is defined as:
\begin{equation}
q(\mathbf{x}_t | \mathbf{x}_{t-1}) = \mathcal{N}(\mathbf{x}_t; \sqrt{1-\beta_t}\mathbf{x}_{t-1}, \beta_t\mathbf{I}),
\end{equation}
where $\beta_t$ and $\mathbf{x}_t$ are the variance schedule controlling the noise level and current data sample with added noise at timestep $t$, respectively. 
The transformer backbone in DiTs processes spatiotemporal tokens through multi-head self-attention layers. Specifically, the DiT backbone utilizes a 3D patch embedding strategy, where the input video is divided into non-overlapping spatiotemporal patches. These patches are then linearly projected and augmented with positional embeddings before being fed into transformer blocks.
The denoising objective is formulated as:
\begin{equation}
\mathcal{L} = \mathbb{E}_{\mathbf{x}_0, \epsilon, t}\left[\|\epsilon - \epsilon_\theta(\mathbf{x}_t, t, \mathbf{c})\|^2\right],
\end{equation}
where $\epsilon$ is the added noise, $\epsilon_\theta$ is the learned denoising network parameterized by $\theta$, and $\mathbf{c}$ represents conditioning information. This formulation allows the model to learn a direct mapping from noisy inputs to predicted noise, which can then be subtracted to recover clean video frames.

\subsection{Synthetic Hair Motion Generation}\label{sec:data_gen}
 While datasets of real subjects with hair motion exist and can be acquired, the low accuracy of hair labeling techniques would hinder the training of diffusion models. Therefore, to construct the hair-specific dataset necessary to train our model, we turn to Computer Graphics (CG) and physics-based simulation~\cite{kugelstadt2016position} and rendering. This approach enables us to create a large corpus of videos with physically plausible hair dynamics, driven by simulated forces, while simultaneously extracting precise, per-pixel ground-truth motion conditions. 
Each data sample in our dataset is a quadruplet $\{\mathbf{I}_{ref}, \mathbf{V}_{gt}, \mathbf{C}_{pose}, \mathbf{C}_{hair}\}$.
\begin{itemize}
    \item $\mathbf{I}_{ref} \in \mathbb{R}^{H \times W \times 3}$ is the static reference image, which serves as the appearance condition. In our setup, this is typically the first frame of the video.
    \item $\mathbf{V}_{gt} \in \mathbb{R}^{T \times H \times W \times 3}$ is the ground truth video sequence, representing the target output. It is rendered using Blender Cycles' physically-based path tracer.
    \item $\mathbf{C}_{pose} \in \mathbb{R}^{T \times H \times W \times 3}$ is the body pose condition, represented as a sequence of camera-space normal renders augmented with a set of 68 facial landmarks. In these renders, the hair geometry is hidden, making the head and body fully visible. This provides dense structural and orientation information for the body, head and face, guiding the overall motion.
    \item $\mathbf{C}_{hair} \in \mathbb{R}^{T \times H \times W \times 3}$ is the hair motion condition, represented as a sequence of renders in which the hair is rasterized as a $U,V,W$ buffer, where the $U,V$ values are the texture coordinates of the scalp at the hair strand root locations and $W$ is the normalized arc-length parameter along the hair curve. It is a dense, per-pixel representation that effectively captures the intricate deformations and flow of hair strands, providing a much richer signal than sparse keypoints or optical flow.
\end{itemize}
This dataset forms the basis for our two-stage training. It provides the explicit signals necessary to train a conditional generation model with motion control that is superior to standard I2V models. Furthermore, our training pipeline (detailed in Section~\ref{sec:domain_lora}) is designed specifically to leverage this data without the realism drawbacks common to 3D-based animation methods.

\subsection{\pose}\label{sec:pose_lora}

The \pose module is our core contribution for motion control. It is a lightweight adapter designed to inject dual motion conditions into the DiT backbone, $\epsilon_\theta$, without altering its pre-trained weights. It features two distinct pathways to process body and hair motion, reflecting the different nature of these signals.

Let the noisy video at timestep $t$ be $\mathbf{x}_t$. This is first encoded by the frozen VAE encoder $\mathcal{E}$ and patchified into spatiotemporal tokens $\mathbf{z}_t = \text{Patchify}(\mathcal{E}(\mathbf{x}_t))$. Our \pose module injects conditions $\mathbf{C}_{pose}$ and $\mathbf{C}_{hair}$ directly into this token space as follows:

\noindent \textbf{Body Motion Pathway:} The body pose condition $\mathbf{C}_{pose}$ (normal map) provides structural guidance. We use a dedicated Pose Encoder, $\mathcal{E}_{pose}$, composed of convolutional layers, as introduced in UniAnimate-DiT~\cite{wang2025unianimate}. $\mathcal{E}_{pose}$ is initialized with the weights of the VAE encoder $\mathcal{E}$ but is fine-tuned during training. The pose condition is encoded and patchified into pose tokens:
\begin{equation}
    \mathbf{z}_{pose} = \text{Patchify}(\mathcal{E}_{pose}(\mathbf{C}_{pose})).
\end{equation}
These pose tokens are injected via element-wise addition to the noisy tokens:
\begin{equation}
    \mathbf{z}_t' = \mathbf{z}_t + \mathbf{z}_{pose}.
\end{equation}
This additive injection modulates the features of the noisy input, effectively biasing the denoising process towards the target pose.

\noindent \textbf{Hair Motion Pathway:} The hair condition $\mathbf{C}_{hair}$ (UVW map) provides dense, fine-grained motion. This signal is processed using the \textbf{\textit{frozen}} VAE encoder $\mathcal{E}$, inspired by DreamActor-M1~\cite{luo2025dreamactor}, and patchified to match the token dimensions:
\begin{equation}
    \mathbf{z}_{hair} = \text{Patchify}(\mathcal{E}(\mathbf{C}_{hair})).
\end{equation}
Unlike the pose condition, the hair tokens are injected by \textit{\textbf{concatenating}} them with the modulated noisy tokens along the sequence length dimension:
\begin{equation}
    \mathbf{z}_{in} = \text{Concat}([\mathbf{z}_t', \mathbf{z}_{hair}]).
\end{equation}
This approach, inspired by unified attention mechanisms~\cite{zhou2024transfusion,luo2025dreamactor,wang2025dreamactor}, allows the DiT's self-attention layers to directly access and utilize the explicit hair motion information as part of the input sequence. The parameters of $\mathcal{E}_{pose}$ and the LoRA weights applied to the DiT's attention blocks constitute the trainable parameters of \pose, denoted $\phi_P$.

\subsection{\domain and Training}\label{sec:domain_lora}

Training $\phi_P$ directly on the CG dataset risks domain overfitting, where the model learns the non-photorealistic style of the simulator. To circumvent this, we propose a two-stage training strategy that separates the learning of domain-specific features from motion-control features.

\noindent \textbf{Stage 1: \domain Pre-training.}
We first pre-train a \domain LoRA module, $\phi_D$, on our synthetic dataset. This stage uses a standard Image-to-Video (I2V) objective, where the model learns to generate video frames $\mathbf{V}_{gt}$ from Reference Image $\mathbf{I}_{ref}$ and text prompts $\mathbf{T}_{prompt}$, describing the content which is captioned by Qwen-2.5-VL~\cite{bai2025qwen2} from $\mathbf{V}_{gt}$. The objective is to capture the general motion dynamics and visual characteristics of the CG domain within $\phi_D$. The loss is:
\begin{equation}
    \mathcal{L}_{domain} = \mathbb{E}_{\mathbf{x}_0 \sim \mathbf{V}_{gt}, \epsilon, t, \mathbf{T}_{prompt}}\left[\|\epsilon - \epsilon_{\theta, \phi_D}(\mathbf{x}_t, t, \mathbf{T}_{prompt})\|^2\right].
\end{equation}
This stage adapts the DiT backbone $\theta$ with LoRA weights $\phi_D$ to the synthetic domain.

\noindent \textbf{Stage 2: \pose Training.}
In the second stage, we \textbf{\textit{freeze}} both the DiT backbone $\theta$ and the pre-trained \domain LoRA $\phi_D$. We then introduce our \pose module $\phi_P$ (which includes the Pose Encoder $\mathcal{E}_{pose}$ and its own LoRA weights). The model is now trained on the full quadruplet $\{\mathbf{I}_{ref}, \mathbf{V}_{gt}, \mathbf{C}_{pose}, \mathbf{C}_{hair}\}$, with $\mathbf{I}_{ref}$ serving as the appearance condition. The objective for $\phi_P$ is to predict the noise $\epsilon$ given the rich conditional inputs:
\begin{equation}
    \mathcal{L}_{pose} = \mathbb{E}_{\mathbf{x}_0 \sim \mathbf{V}_{gt}, \epsilon, t, \mathbf{c}}\left[\|\epsilon - \epsilon_{\theta, \phi_D, \phi_P}(\mathbf{x}_t, t, \mathbf{c})\|^2\right],
\end{equation}
where $\mathbf{c} = \{\mathbf{I}_{ref}, \mathbf{C}_{pose}, \mathbf{C}_{hair}\}$. Because $\phi_D$ is frozen, it provides a stable, CG-domain-adapted feature space, forcing $\phi_P$ to focus exclusively on learning the mapping from the motion conditions ($\mathbf{C}_{pose}$, $\mathbf{C}_{hair}$) to the desired output.

\noindent \textbf{Inference.}
At inference time, we \textbf{\textit{discard}} the \domain LoRA $\phi_D$ entirely. Inference is performed using only the original DiT backbone $\theta$ and our trained \pose module $\phi_P$:
\begin{equation}
    \epsilon_{pred} = \epsilon_{\theta, \phi_P}(\mathbf{x}_t, t, \mathbf{c}).
\end{equation}
This strategy allows \ourmodel to leverage the precise, fine-grained motion control learned by $\phi_P$ from the synthetic data, while completely shedding the non-photorealistic domain artifacts captured by $\phi_D$. The final generation is thus guided by $\phi_P$ but rendered using the original, photorealistic generative priors of the DiT backbone.

\begin{figure*}[h]\vspace{-5pt}
\centering
 \includegraphics[width=\linewidth]{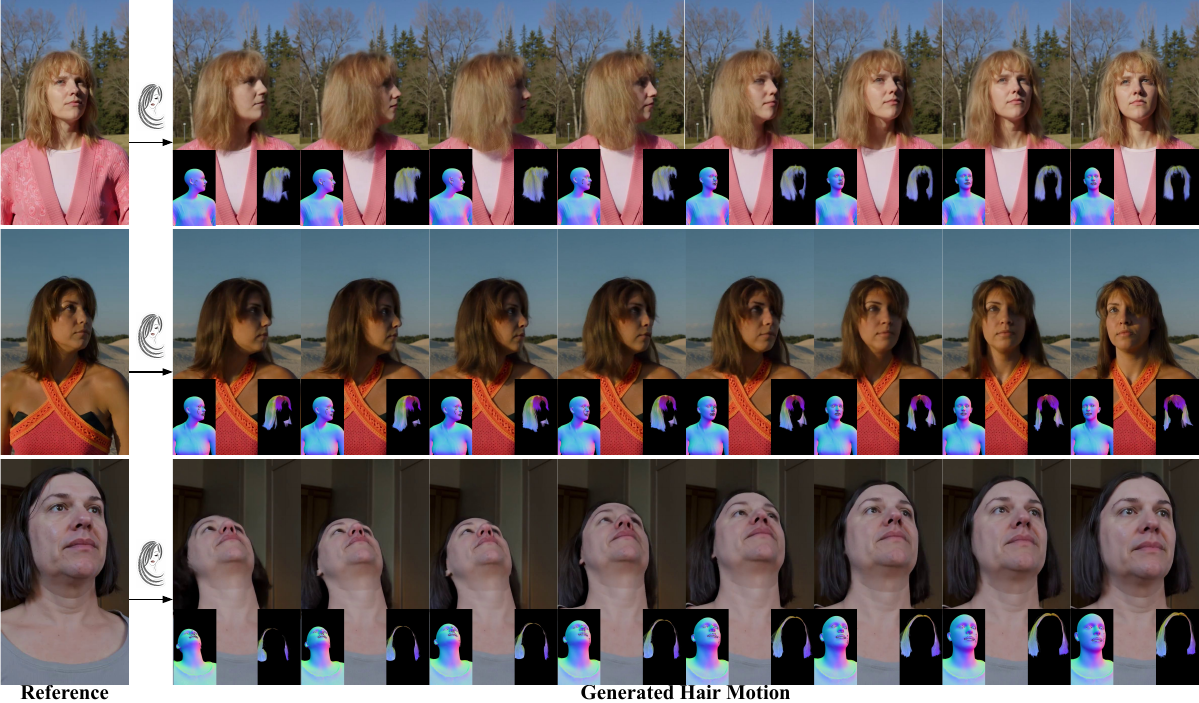}
  \vspace{-15pt}
   \caption{Photorealistic motions generated by HairWeaver. The reference images are photorealistic human subjects generated by Flux~\cite{flux2024}.}
   \vspace{-20pt}
    \label{fig:flux}
\end{figure*}

\section{Experiments}
\label{exp}

\subsection{Implementation Details}\label{implement}

\noindent \textbf{Dataset}\label{data} Our model is trained on a synthetic human video dataset, which contains 83 minutes of synthetic videos (1,500 videos of 100 frames each with FPS=30) generated from the CG simulator. This dataset was specifically created to provide clean examples of complex hair dynamics, which are critical for our task. All videos and conditions were processed to a resolution of 896 (height) $\times$ 512 (width). This curated dataset ensures our model is exposed to a broad distribution of high-quality motions and expressive details. We test our model on two test sets.
We first used the model trained on the synthetic data and tested it on the self-collected hair motion test-set, which is also generated from the simulator but with different (synthetic) human subjects than those in the training data. This test set comprises 10 randomly selected videos featuring diverse hair and body motions. 
% To further validate the controllable hair motion generation quality of our model in photorealistic cases, we then utilize Flux~\cite{flux2024} to generate 1k different photorealistic human reference images and use the trained model to generate 1k videos with hair and body motions. We further combine these videos with 350 videos from TikTok~\cite{Jafarian_2021_CVPR_TikTok} dataset, which was widely used by previous work~\cite{chang2023magicpose,wang2023disco,xu2024magicanimate}, as training data and 
We also evaluate the result on the NeRSemble~\cite{kirschstein2023nersemble} test-set. This test set comprises 30 videos featuring diverse photorealistic hair and body motions with human subjects facing the camera viewpoint, which are randomly sampled from the hair subset of the whole NeRSemble v2 Dataset. 
Note that since UVW maps and body normal maps are unavailable for the NeRSemble dataset, we fine-tuned a \pose using the alpha map as the hair condition and densepose map~\cite{guler2018densepose} as the body condition in this case. 
The purpose of such an experiment is to demonstrate our framework's flexibility: while it achieves peak precision with UVW maps in CG environments, it can successfully utilize coarser real-world signals while retaining the physical priors learned during synthetic pre-training.

\noindent \textbf{Model Training and Inference}\label{model}
Our framework is built upon the LTX-Video-0.9.8~\cite{HaCohen2024LTXVideo}, whose DiT backbone weights remain frozen during all training phases. For initialization, both the pose encoder for body and the VAE-Encoder for hair inherit the weights from the pretrained LTX-Video VAE-Encoder, while the layers in \pose and \domain are initialized with zeros.
We employ a two-stage training strategy. In the first stage, we train only the \domain for 10,000 steps in a standard image-to-video manner with reference image and text prompt captioned by Qwen-2.5-VL~\cite{bai2025qwen2} as input. In the second stage, we freeze the \domain and train the \pose and Pose Encoder for 10,000 steps on our primary task of motion-conditioned image-to-video animation.
For all training, we use the AdamW optimizer with a learning rate of $2e^{-4}$. The model is trained on video clips of 97 frames. All experiments were conducted on 8 NVIDIA H200 GPUs with a batch size of 8.

\begin{table*}[ht]
\centering
\caption{Comparison to state-of-the-art video animation and video-to-video translation methods on our self-collected hair motion test-set.}\vspace{-10pt}
\label{tab:cg_test_set}
\resizebox{\textwidth}{!}{
\begin{tabular}{l|ccccc|ccccc|c}
\toprule
\multirow{2}{*}{\textbf{Method}} & \multicolumn{5}{c|}{\textbf{Hair}} & \multicolumn{5}{c|}{\textbf{Full Body}} & \multirow{2}{*}{\textbf{Avg Infer Time(s)$\downarrow$}} \\
\cmidrule(lr){2-6} \cmidrule(lr){7-11}
& \textbf{SSIM$\uparrow$} & \textbf{PSNR$\uparrow$} & \textbf{LPIPS$\downarrow$} & \textbf{FID$\downarrow$} & \textbf{cd-FVD$\downarrow$} & \textbf{SSIM$\uparrow$} & \textbf{PSNR$\uparrow$} & \textbf{LPIPS$\downarrow$} & \textbf{FID$\downarrow$} & \textbf{cd-FVD$\downarrow$} & \\
\midrule
LTX-Video-0.9.8-13B~\cite{HaCohen2024LTXVideo} & 0.9642 & 30.3565 & 0.0468 & 156.8218 & 827.5890 & 0.7885 & 19.9854 & 0.2751 & 113.9513 & 866.8544 & \textbf{56} \\
Wan-2.2-14B~\cite{wan2025} & 0.9153 & 24.8174 & 0.1326 & 172.6361 & 802.6791 & 0.5535 & 12.7273 & 0.6345 & 89.8836 & 868.1351 & 476 \\
LTX-Video-ICLora~\cite{HaCohen2024LTXVideo}& 0.9700 & 30.6511 & 0.0365 & 108.1872 & 652.3275 & 0.8665 & 24.1449 & 0.1446 & 45.4541 & 423.4040 & \underline{58} \\

UniAnimate-DiT~\cite{wang2025unianimate} & \underline{0.9761} & \underline{35.4174} & 0.0304 & 66.1997 & 678.9257 & \underline{0.8724} & 25.6764 & \underline{0.1407} & \underline{45.0817} & 448.8855 & 870 \\

Wan-2.2-Animate-14B~\cite{wan2025} & 0.9758 & 35.3174 & \underline{0.0299} & 59.6534 & 587.4605 & 0.8400 & 25.2877 & 0.1803 & 49.4041 & 461.6013 & 312 \\

Wan-VACE~\cite{vace} + Hair + Body & 0.9724 & 34.6198 & 0.0348 & \underline{57.9023} & \underline{463.2543} & 0.8514 & \underline{26.4512} & 0.1547 & 51.3914 & \underline{412.5809} & 314 \\

\ourmodel & \textbf{0.9794} & \textbf{37.6347} & \textbf{0.0233} & \textbf{50.5938} & \textbf{434.1582} & \textbf{0.8948} & \textbf{27.7903} & \textbf{0.1127} & \textbf{43.2786} & \textbf{407.1929} & 62 \\

\bottomrule
\end{tabular}
}
\vspace{-5pt}
\end{table*}
\begin{table*}[ht]
\centering
\caption{Comparison to state-of-the-art video animation methods on NeRSemble~\cite{kirschstein2023nersemble} test-set.}\vspace{-10pt}
\label{tab:ns_test_set}
\resizebox{\textwidth}{!}{
\begin{tabular}{l|ccccc|ccccc}
\toprule
\multirow{2}{*}{\textbf{Method}} & \multicolumn{5}{c|}{\textbf{Hair}} & \multicolumn{5}{c}{\textbf{Portrait}} \\
\cmidrule(lr){2-6} \cmidrule(lr){7-11}
& \textbf{SSIM$\uparrow$} & \textbf{PSNR$\uparrow$} & \textbf{LPIPS$\downarrow$} & \textbf{FID$\downarrow$} & \textbf{cd-FVD$\downarrow$} & \textbf{SSIM$\uparrow$} & \textbf{PSNR$\uparrow$} & \textbf{LPIPS$\downarrow$} & \textbf{FID$\downarrow$} & \textbf{cd-FVD$\downarrow$} \\
\midrule
LTX-Video-0.9.8-13B~\cite{HaCohen2024LTXVideo} & 0.9131 & 23.40 & 0.1020 & 120.74 & 891.18 & 0.6871 & 17.83 & 0.3878 & 73.50 & 703.35 \\
Wan-2.2-14B~\cite{wan2025} & 0.9150 & 21.65 & 0.1051 & 113.11 & 652.18 & 0.6865 & 17.28 & 0.3830 & 51.43 & 503.07 \\
LTX-Video-ICLora~\cite{HaCohen2024LTXVideo} & 0.9397 & 24.95 & 0.0710 & 44.11& 384.41 & 0.7635 &20.99 &0.2683 &28.42 & \underline{277.77} \\
UniAnimate-DiT~\cite{wang2025unianimate} & 0.9350 & 25.33 & 0.0810 & 66.29 & 477.13 & 0.7545 & 20.32 & 0.3059 & 52.44 & 379.12 \\
Wan-2.2-Animate-14B~\cite{wan2025} & \underline{0.9544} & \underline{28.75} & \underline{0.0599} & \underline{20.48} & \underline{328.67} & \underline{0.8011} & \underline{23.62} & \underline{0.2030} & \underline{24.32} & {314.65} \\
\ourmodel & \textbf{0.9670} & \textbf{34.34} & \textbf{0.0477} & \textbf{17.79} & \textbf{286.25} & \textbf{0.8291} & \textbf{26.47} & \textbf{0.1763} & \textbf{19.43} & \textbf{212.61} \\
\bottomrule
\end{tabular}
}
\vspace{-5pt}
\end{table*}
\subsection{Evaluations and Comparisons}
\label{comp}
\noindent \textbf{Metrics.} 
We evaluate our model against state-of-the-art methods using a suite of standard metrics that assess different aspects of generation quality.

\begin{table*}[ht]
\centering
\caption{Ablation Analysis. \textbf{DiT+Pose Encoder} denotes the DiT backbone is trained with Pose Encoder only without \domain and \pose. \textbf{DiT+Pose-IC-LoRA} denotes the DiT backbone is trained with In-Context-Lora, the same as LTX-Video-ICLora~\cite{HaCohen2024LTXVideo}, without \domain and \pose. \textbf{w/o \domain} denotes the \domain module is removed from the \ourmodel pipeline. \textbf{Wan-VACE+UVW Map} denotes we finetune Wan-VACE with the same UVW Map as \ourmodel. \textbf{HairWeaver+Alpha Map} denotes we replace the UVW Map with Alpha Map as hair condition and replace the normal map with DensePose as body condition.} 
\vspace{-10pt}
\label{tab:cg_test_ablation}
\resizebox{\textwidth}{!}{
\begin{tabular}{l|ccccc|ccccc}
\toprule
\multirow{2}{*}{\textbf{Method}} & \multicolumn{5}{c|}{\textbf{Hair}} & \multicolumn{5}{c}{\textbf{Full Body}} \\
\cmidrule(lr){2-6} \cmidrule(lr){7-11}
& \textbf{SSIM$\uparrow$} & \textbf{PSNR$\uparrow$} & \textbf{LPIPS$\downarrow$} & \textbf{FID$\downarrow$} & \textbf{cd-FVD$\downarrow$} & \textbf{SSIM$\uparrow$} & \textbf{PSNR$\uparrow$} & \textbf{LPIPS$\downarrow$} & \textbf{FID$\downarrow$} & \textbf{cd-FVD$\downarrow$} \\
\midrule
DiT+Pose Encoder & 0.9158 & 27.4486 & 0.1214 & 94.0048 & 654.8677 & 0.5623 & 12.9256 & 0.6481 & 91.1008 & 431.4920 \\
DiT+Pose-IC-LoRA & 0.9700 & 30.6511 & 0.0365 & 108.1872 & 652.3275 & 0.8665 & 24.1449 & 0.1446 & 45.4541 & 423.4040 \\
w/o \domain & 0.9693 & 36.7183 & \underline{0.0236} & \textbf{49.0808} & 447.7628 & \underline{0.8828} & 26.5544 & \underline{0.1184} & \textbf{38.4230} & 416.9271 \\
Wan-VACE+UVW Map & \underline{0.9724} & 34.6198 & 0.0348 & 57.9023 & 463.2543 & 0.8514 & 26.4512 & 0.1547 & 51.3914 & 412.5809 \\
HairWeaver+Alpha Map & 0.9693 & \underline{36.7297} & \underline{0.0236} & \underline{49.0856} & \underline{436.9389} & 0.8710 & \underline{26.9807} & 0.1340 & 46.1248 & \underline{410.2243} \\
\ourmodel & \textbf{0.9794} & \textbf{37.6347} & \textbf{0.0233} & {50.5938} & \textbf{434.1582} & \textbf{0.8948} & \textbf{27.7903} & \textbf{0.1127} & \underline{43.2786} & \textbf{407.1929} \\
\bottomrule
\end{tabular}
}
\vspace{-16pt}
\end{table*}

\begin{itemize}
    \item \textbf{Reconstruction and Perceptual Quality:} We measure frame-level similarity to the ground truth using Peak Signal-to-Noise Ratio (\textbf{PSNR}), Structural Similarity Index Measure (\textbf{SSIM}), \textbf{L1} distance, and the Learned Perceptual Image Patch Similarity (\textbf{LPIPS})~\cite{zhang2018unreasonable}.
    
    \item \textbf{Video Realism and Temporal Consistency:} To evaluate the overall quality and realism of the generated video distribution, we report the Fréchet Inception Distance (\textbf{FID}) and the Content-Debiased Fréchet Video Distance (\textbf{cd-FVD})~\cite{ge2024content}. The cd-FVD metric provides a more robust measure of temporal quality than the standard FVD~\cite{unterthiner2018towards} by disentangling content from motion.
    
\end{itemize}

\noindent These metrics are widely adopted in recent human video animation research~\cite{wang2023disco,xu2024magicanimate,chang2023magicpose,zhang2024mimicmotion}, providing a standardized basis for comparison.

% \begin{table}[htp]
% \centering
% \caption{30 participants used a forced-choice rating to select videos generated by six methods from eight identities from test set. (Full table presented in the supplementary).\ourmodel generates the most realistic human videos with hair motions.
% }
% \vspace{-7pt}
% \label{tab:user_study}
% \scalebox{0.95}
% {\begin{tabular}{lcccccccccccc}
% \toprule
% Method  & {\textbf{Average}}\\
% \midrule
% LTX-Video-0.9.8-13B~\cite{HaCohen2024LTXVideo} & 6.9\%\\

% Wan-2.2-14B~\cite{wan2025} & 8.2\%\\
% LTX-Video-ICLora~\cite{HaCohen2024LTXVideo} & 9.0\%\\
% UniAnimate-DiT~\cite{wang2025unianimate} & 9.5\%\\
% Wan-2.2-Animate-14B~\cite{wan2025} & \underline{16.4\%}\\
% \ourmodel & \textbf{49.9\%}\\
% \bottomrule
% \end{tabular}}
% % \vspace{-10pt}
% \end{table}

\begin{wraptable}{l}{6.5cm} % 'r' for right-side alignment, 6.5cm for width
% \vspace{-15pt}
    \centering
    \vspace{-15pt}
    \caption{30 participants used a forced-choice rating to select videos generated by six methods from eight identities. \ourmodel generates the most realistic videos.}
    \label{tab:user_study}
    % \vspace{-5pt} % Adjust spacing between caption and table
    \scalebox{0.95}{
        \begin{tabular}{lc} % Simplified to 2 columns based on your data
            \toprule
            Method  & \textbf{Average} \\
            \midrule
            LTX-Video-0.9.8-13B~\cite{HaCohen2024LTXVideo} & 6.9\% \\
            Wan-2.2-14B~\cite{wan2025} & 8.2\% \\
            LTX-Video-ICLora~\cite{HaCohen2024LTXVideo} & 9.0\% \\
            UniAnimate-DiT~\cite{wang2025unianimate} & 9.5\% \\
            Wan-2.2-Animate-14B~\cite{wan2025} & \underline{16.4\%} \\
            \ourmodel & \textbf{49.9\%} \\
            \bottomrule
        \end{tabular}
    }
    \vspace{-17pt} % Adjust spacing below the table
\end{wraptable}

\noindent \textbf{Quantitative Comparison}\label{quantitative}
We benchmark our method, \ourmodel, against several state-of-the-art video generation models that represent distinct architectural approaches. These baselines include: (1) Image-to-Video methods, Wan2.2~\cite{wan2025} and LTX-Video-0.9.8~\cite{HaCohen2024LTXVideo}, and (2) Human Video Animation methods with additional control, such as LTX-Video-0.9.8-ICLora~\cite{HaCohen2024LTXVideo}, UniAnimate-DiT~\cite{wang2025unianimate}, and Wan2.2-Animate~\cite{wan2025}.
Our primary goal is to improve the quality of dynamic hair generation. While our model is able to leverage a driving signal for the hair that is unavailable to other models, we still find it relevant to conduct quantitative evaluations against those models to 1) validate that, regardless of control signals available or not, our model is the only one able to get close to ground truth and 2) to set a baseline for future research. Our quantitative results are presented in Table~\ref{tab:cg_test_set}, where we compare performance on the self-collected hair motion CG test-set. Note that we use the hair segmentation mask from Matte-Anything~\cite{yao2024matte} to segment the videos and calculate the metrics for hair area. The results demonstrate that \ourmodel~consistently and significantly outperforms all baseline models across the relevant metrics. This indicates that our proposed method successfully generates animations with more vivid and expressive dynamics. We also compare the efficiency of \ourmodel to previous methods by reporting average inference time (in seconds) per video sample. Our method is more efficient compared to state-of-the-art animation methods~\cite{wang2025unianimate,wan2025}. As mentioned in Section~\ref{implement}, we also test the model on the photorealistic NeRSemble~\cite{kirschstein2023nersemble} test-set. The result is presented in Table~\ref{tab:ns_test_set}. We observe that the hair motion quality of \ourmodel~outperforms all baselines across the relevant metrics as well.

\begin{wrapfigure}{r}{0.5\textwidth}
    \centering
    \vspace{-35pt} % Adjust to pull the figure up if there's too much white space
    \includegraphics[width=0.48\textwidth]{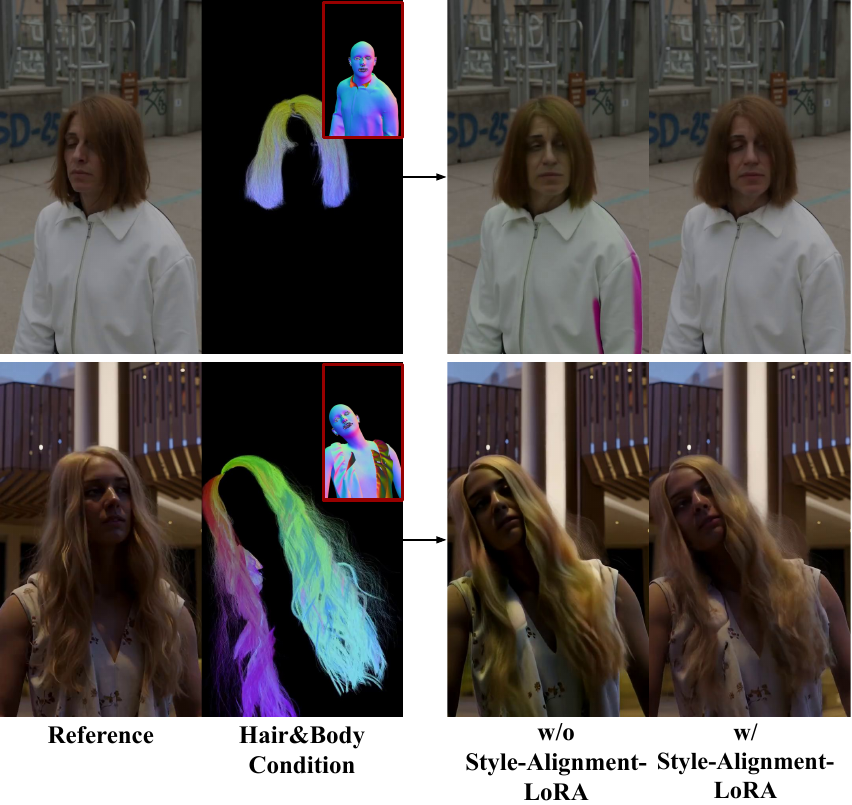}
    \vspace{-10pt}
    \caption{Ablation analysis of \domain. We visualize the generation without (w/o) and with (w/) \domain. The one w/o such a module cannot preserve reference's appearance when it's a photorealistic image.}
    \label{fig:ablation}
    \vspace{-80pt} % Adjust to pull the text up below the caption
\end{wrapfigure}

% \begin{figure*}[t!]
% \vspace{-5pt}
% \centering
%  \includegraphics[width=0.98\linewidth]{Figures/Exp_tiktok_2.png}
%   % \vspace{-15pt}
%    \caption{{Animation generated by our \ourmodel, from left-to-right: Reference Image / Pose Condition / Baseline / Ours / Driving Video.}
%    }
%    \vspace{-5pt}
%     \label{fig:tiktok_2}
% \end{figure*}

% \begin{figure*}[t!]
% \vspace{-5pt}
% \centering
%  \includegraphics[width=0.98\linewidth]{Figures/Exp_ood_2.png}
%   % \vspace{-15pt}
%    \caption{{Animation generated by our \ourmodel, from left-to-right: Reference Image / Pose Condition / Baseline / Ours / Driving Video.}
%    }
%    \vspace{-5pt}
%     \label{fig:ood_2}
% \end{figure*}

\begin{wrapfigure}{r}{0.5\textwidth}
  \vspace{-15pt}
    \centering
\includegraphics[width=0.48\textwidth]{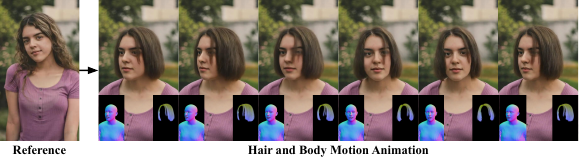}
  \vspace{-5pt}
\captionof{figure}{Motions generated by HairWeaver. The reference image is collected from in-the-wild photoreal source.}
   \vspace{-15pt}
\label{fig:rebuttal_real}
\end{wrapfigure}

\begin{figure*}[t!]\vspace{-5pt}
\centering
 \includegraphics[width=0.99\linewidth]{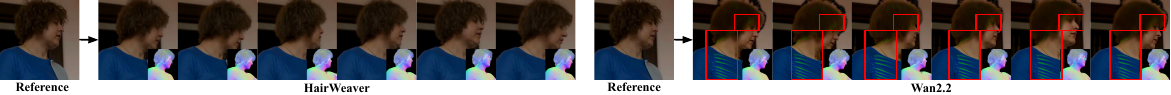}
  \vspace{-5pt}
   \caption{{Comparison to an image-conditioned video-to-video method, Wan-VACE~\cite{wan2025}, with same hair and body conditions as HairWeaver.
   }
   }
   \vspace{-5pt}
    \label{fig:wan_vace_comp}
\end{figure*}

\noindent \textbf{Qualitative Comparison}\label{qualitative}
We qualitatively compare the dynamic hair and body motion generation of \ourmodel on NeRSemble test-set with the animation baselines in Figure~\ref{fig:ns_comp}.
We also provide a comparison to an image-conditioned video-to-video method, Wan-VACE~\cite{vace}, with the same conditions as ours in Table~\ref{tab:cg_test_set}. This demonstrates that our architectural contributions (Motion-Context-LoRA + Sim2Real strategy) provide genuine improvement beyond simply having richer conditioning.
Visual comparison in Figure~\ref{fig:wan_vace_comp} shows that Wan-VACE, struggles to maintain identity consistency while translating fine-grained hair details into realistic motions.\\
We also provide more photorealistic visualizations in Figure.~\ref{fig:flux}, where the reference images are generated by Flux~\cite{labs2025flux1kontextflowmatching}, and in Figure~\ref{fig:rebuttal_real} where the reference is real-world image. The photorealistic quality depends on the realism of the reference image; with authentic photographs, our model maintains photorealism through the \domain, which prevents CG artifacts from leaking into the generation. More video visualizations are presented in the project page of the supplementary materials.

\noindent \textbf{User Study}\label{user}
 We conducted a user study to compare \ourmodel with previous work~\cite{HaCohen2024LTXVideo,wan2025,wang2025unianimate}. We collect reference images, ground truth videos, and animation results from previous works and \ourmodel for 8 video subjects. For each subject, we visualize the reference, animation results, and ground truth videos side by side, with the animation results anonymized and randomly ordered. We ask 50 users to choose the \textbf{best} methods according to the following two criteria: (1) follows the hair motions and head poses in the Real Video and (2) preserves the identity and appearance of the reference image (person). We present the result in Table~\ref{tab:user_study}. We observe that the users prefered \ourmodel more than baseline methods. For more details, please refer to the supplementary material.
We also provide more video visualizations in our supplementary website for more results with long hair. These videos can also be directly viewed in subfolder  "hairweaver\_page/static/flux\_more\_viz\_compressed", e.g., 011.mp4; 027.mp4; 039.mp4; 043.mp4; 044.mp4.

\noindent \textbf{Ablation Analysis}\label{ablation}
We provide the ablation analysis in Table~\ref{tab:cg_test_ablation}. Since the self-collected test set has ground-truth video from the CG simulator domain, it is challenging to observe the improvement from \domain in overcoming the domain gap between the CG training data and the photorealistic test data. To this end, we present a visualization of the effectiveness of such a module in Figure.~\ref{fig:ablation}. The proposed \domain ability to preserve appearance is evident.
To quantify the benefit of richer conditioning, we also provide a comparison on the CG test set between the UVW-conditioned model and the alpha-conditioned variant.
This confirms that UVW maps provide superior fine-grained control, while alpha maps still produce results that significantly outperform all baselines — validating that simulation-quality dynamics are learned and expressed even from coarse conditioning.

\subsection{End-to-end practical application}\label{sec:practical}

% \section{Details of Photoreal Inference}

In this section, we provide the practical details of our pipeline for generating  high quality photorealistic animations with fine-grained control from CG animations.
Given a CG animation, we render it using the same simulation and rendering pipeline that 
was employed previously for generating the CG training dataset. Specifically, we use Blender Cycles' physically-based 
path-tracer to render 1) the shaded images $\mathbf{V}_{gt}$ and 2) the normal and \textit{UVW} images ($\mathbf{C}_{pose}$ and $\mathbf{C}_{hair}$ respectively) required for guiding the generation.
We then take the first shaded frame and transform it into a more photorealistic reference image using an image-to-image pipeline~\cite{meng2022sdedit} based on Flux~\cite{flux2024} --- the resulting image becoming our $\mathbf{I}_{ref}$.
Since it is desirable for our image-to-image process to preserve the alignment between the source and the result as much as possible, we add only a small amount of noise to the source image and we provide a detailed 
prompt which thoroughly describes it.
In our implementation, we set the noising timestep to 0.35 and perform 100 denoising steps following the Euler integration scheme. We use the Qwen-2.5 Vision Language Model~\cite{bai2025qwen2} to generate the detailed prompt from the source image.\\
In this case, our approach and model are immediately useful for sim2real applications, e.g. human-centric synthetic data generation. UVW maps and body normal maps are readily available from the CG simulator and the input reference can be stylized as photorealistic identities using the flux-based approach, e.g., Figure.~\ref{fig:flux} illustrates how to achieve the photorealism enhancement. With such synthetic data, users can train models for other downstream tasks, e.g. creating digital avatars with punctual hairstyle editing or try-on. 

\subsection{Limitations and Future Works}
\label{limitation}

The model's performance can degrade in scenarios with extreme deviations between the driving pose and the reference image. For instance, when the driving video involves significant zooming that leads to large changes in scale, the preservation of the subject's appearance and identity can be compromised. Furthermore, like many generative models, our method struggles to render certain details, particularly when generating consistently accurate and realistic hands. We attribute these challenges primarily to the scope of our training data and the limitations of the backbone model's (LTX-Video~\cite{HaCohen2024LTXVideo}) generation ability. Future work could address these by incorporating more diverse training data and employing more advanced backbone models.

\section{Conclusion}
\label{conclusion}
In this paper, we introduce \textbf{\ourmodel}, a novel diffusion-based framework designed to address a critical limitation in human video animation: the synthesis of expressive and realistic hair motion. We identified that existing methods, while proficient at transferring body pose, often fail to model secondary dynamics, resulting in characters with static, unnatural hair that undermines realism.
Our solution leverages a pre-trained video DiT, enhanced with two key components. The \textbf{\pose} seamlessly integrates hair motion control by adding additional context to the input latent, effectively guiding the animation while preserving the rich generative priors of the backbone model. Concurrently, the \textbf{\domain} improves the model's flexibility, enabling generalization to diverse photorealistic identities. By training on a synthetic dataset with simulated videos rich in hair dynamics, our model learns to generate motion that is not only temporally coherent but also physically plausible.

\section{Ethics Statement}
\label{ethics}
We clarify that, except for those in the NeRSemble~\cite{kirschstein2023nersemble} dataset, all characters in this paper are fictional. We strongly condemn any misuse of generative artificial intelligence that could harm individuals or disseminate misinformation. While we acknowledge the potential for misuse in human-centered animation generation, we are dedicated to upholding the highest ethical standards in our research. This commitment includes strict adherence to legal frameworks, respect for privacy, and a focus on promoting the generation of positive and constructive content. We have submitted the legal review for the usage of such data in our work and it's still pending approval from the legal team of our organization.

\section*{Acknowledgement}
The authors would like to thank Michelle Hill, Lindsey Pollock, Jennie Antonio, and Steffi Liem for their help setting up and operating the motion capture sessions, and Nicky He for processing the motion capture data. We also thank Priyamvad Ravindra Deshmukh and the Meta Metasim team for providing the digital human asset library and pipeline.
Soleymani's work was sponsored by the Army Research Office and was accomplished under Cooperative Agreement Number W911NF-25-2-0040. The views and conclusions contained in this document are those of the authors and should not be interpreted as representing the official policies, either expressed or implied, of the Army Research Office or the U.S. Government. The U.S. Government is authorized to reproduce and distribute reprints for Government purposes notwithstanding any copyright notation herein.

\clearpage
\newpage

% \input{sec/X_suppl}
% \clearpage
% \newpage

\bibliographystyle{splncs04}
\bibliography{main}

\end{document}

% ---------------------------------------------------------------
% TODO REVIEW: Replace with your title
\title{Supplementary Materials - \ourmodel\includegraphics[height=1.2em]{Figures/Emoji/hair_emoji.png}: Few-Shot Photorealistic Hair Motion Synthesis with Sim-to-Real Guided Video Diffusion}

% TODO REVIEW: If the paper title is too long for the running head, you can set
% an abbreviated paper title here. If not, comment out.
\titlerunning{\ourmodel}

% TODO FINAL: Replace with your author list. 
% Include the authors' OCRID for the camera-ready version, if at all possible.
\author{Di Chang\inst{1,2}\orcidlink{0009-0002-0281-8896} \and
Ji Hou\inst{1}\orcidlink{0000-0002-5244-8953} \and
Aljaz Bozic\inst{1}\orcidlink{0009-0002-2985-6921} \and
Assaf Neuberger\inst{1}\and \\
Felix Juefei-Xu\inst{1}\orcidlink{0000-0002-0857-8611} \and
Olivier Maury\inst{1}\orcidlink{0009-0004-3985-1934} \and
Gene Wei-Chin Lin\inst{1} \and
Tuur Stuyck\inst{1}\orcidlink{0000-0003-1892-2137} \and
Doug Roble\inst{1}\orcidlink{0009-0004-3415-4283} \and
Mohammad Soleymani\inst{2}\orcidlink{0000-0002-5873-1434} \and
Stephane Grabli\inst{1}\orcidlink{0009-0006-5616-1753}}

% TODO FINAL: Replace with an abbreviated list of authors.
\authorrunning{Di Chang et al.}
% First names are abbreviated in the running head.
% If there are more than two authors, 'et al.' is used.

% TODO FINAL: Replace with your institution list.
\institute{Meta \and University of Southern California\\
\url{https://boese0601.github.io/hairweaver/} \\
\email{dichang@usc.edu}}

\maketitle

\setcounter{page}{1}

\section{Details of User Study}
\label{sec:detail_user}
\begin{table*}[htp]
\centering
\vspace{-20pt}
\caption{The user study with 30 participants. We collect the number of votes for eight video subjects from test set by six methods and report the percentage. Our \ourmodel generates the most realistic human videos with hair motions.
}
\vspace{-10pt}
\label{tab:user_study_full}
\scalebox{0.69}
{\begin{tabular}{lcccccccccccc}
\toprule
Method  & {\textbf{Subject1}} & {\textbf{Subject2}}& {\textbf{Subject3}}& {\textbf{Subject4}}&{\textbf{Subject5}} & {\textbf{Subject6}} & {\textbf{Subject7}}& {\textbf{Subject8}}& {\textbf{Average}}\\
\midrule
LTX-Video-0.9.8-13B~\cite{HaCohen2024LTXVideo} & 0.0\% & 6.9\% & 10.3\% & 10.3\% & 6.9\% & {10.3\%} & 6.9\% & 3.4\% & 6.9\%\\
Wan-2.2-14B~\cite{wan2025} & 10.3\% & 3.4\% & 10.3\% & 10.3\% & 3.4\% & 6.9\% & 6.9\% & 13.8\% & 8.2\%\\
LTX-Video-ICLora~\cite{HaCohen2024LTXVideo} & \underline{20.7\%} & 6.9\% & 6.9\% & 10.3\% & \underline{10.3\%} & 6.9\% & 6.9\% & 3.4\% & 9.0\%\\
UniAnimate-DiT~\cite{wang2025unianimate} & 10.3\% & 10.3\% & 10.3\% & 10.3\% & 6.9\% & 6.9\% & \underline{17.2\%} & 3.4\% & 9.5\%\\
Wan-2.2-Animate-14B~\cite{wan2025} & 10.3\% & \underline{24.1\%} & \underline{17.2\%} & \underline{24.1\%} & 6.9\% & \underline{13.8\%} & 6.9\% & \underline{27.6\%} & \underline{16.4\%}\\
\ourmodel & \textbf{48.3\%} & \textbf{48.3\%} & \textbf{41.4\%} & \textbf{34.5\%} & \textbf{65.5\%} & \textbf{55.2\%} & \textbf{55.2\%} & \textbf{48.3\%} & \textbf{49.9\%}\\
\bottomrule
\end{tabular}}
\vspace{-10pt}
\end{table*}
In this section, we provide a comprehensive user study for qualitative comparison between \ourmodel and previous works~\cite{HaCohen2024LTXVideo,wang2025unianimate,wan2025}. 
We generate 8 different human animation results from all baseline models and \ourmodel, where the results are anonymized and shuffled.
On the online platform \textbf{Prolific}, we ask 30 users to choose the only one \textbf{best} method from all videos for each animation result.

\subsection{Results and Statistical Analysis:}
We present the full user study result in Table.~\ref{tab:user_study_full}. % Statistical Analysis Section
To validate the user study results, we conduct a chi-square test of independence~\cite{mchugh2013chi} to determine whether the preference distribution across methods is statistically significant. The null hypothesis ($H_0$) states that there is no significant difference in user preferences among the six methods.

\noindent \textbf{Chi-Square Test Results.} 
We construct a $6 \times 8$ contingency table with vote counts for six methods across eight subjects. The chi-square test yields $\chi^2 = 122.47$, with 35 degrees of freedom and $p < 0.001$, strongly rejecting the null hypothesis. This indicates that user preferences are significantly different across methods at the $\alpha = 0.05$ significance level.

\noindent \textbf{Pairwise Comparisons.}
We further conduct pairwise chi-square tests between \ourmodel and each baseline method. Table~\ref{tab:pairwise_stats} shows the results. All comparisons demonstrate highly significant differences ($p < 0.001$), confirming that \ourmodel is significantly preferred over all baseline methods.

\noindent \textbf{Effect Size.}
We compute Cramér's V as a measure of effect size, yielding $V = 0.361$, indicating a large effect. This suggests that the observed differences in user preferences are not only statistically significant but also practically meaningful.

\noindent \textbf{Conclusion.}
The statistical analysis strongly supports our claim that \ourmodel generates more realistic human videos with hair motions compared to existing methods. With an average preference rate of 49.9\%, our method receives approximately 3 times more votes than the best baseline (Wan-2.2-Animate-14B at 16.4\%) and significantly outperforms all competing methods ($p < 0.001$).

\begin{figure}[t!]\vspace{-5pt}
\centering
 \includegraphics[width=0.99\linewidth]{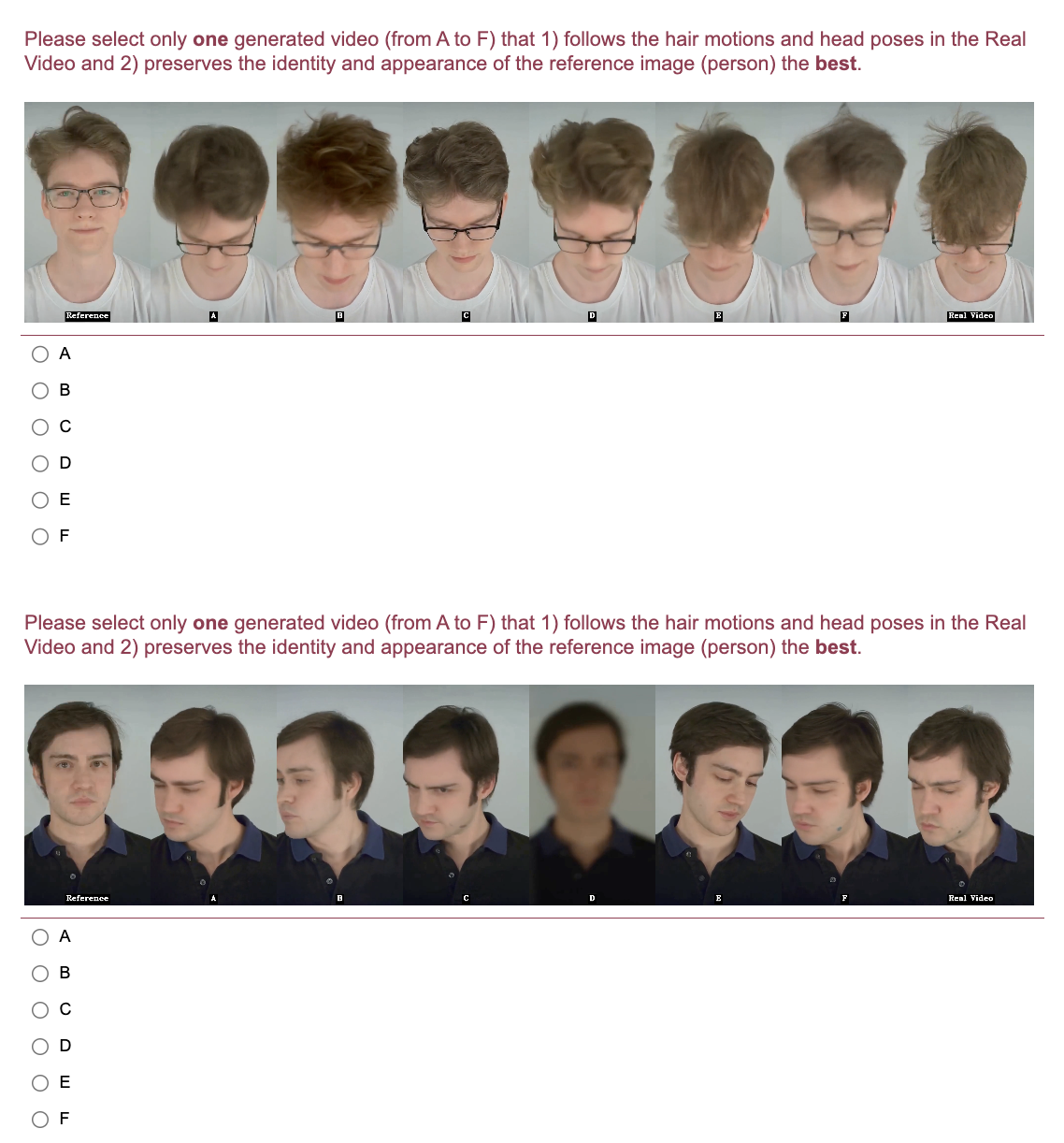}
  % \vspace{-15pt}
   \caption{{Screen shot of the user study.
   }
   }
   % \vspace{-12pt}
    \label{fig:user}
\end{figure}

% Optional: Pairwise comparison table
\begin{table}[h]
\centering
\caption{Pairwise chi-square tests between \ourmodel and baseline methods. All comparisons show highly significant differences ($***$ indicates $p < 0.001$).}
\label{tab:pairwise_stats}
\scalebox{0.75}{
\begin{tabular}{lcccc}
\toprule
\textbf{Comparison} & $\chi^2$ & \textbf{DoF} & $p$\textbf{-value} & \textbf{Sig.} \\
\midrule
\ourmodel vs. LTX-Video-0.9.8-13B & 87.03 & 7 & $< 0.001$ & *** \\
\ourmodel vs. Wan-2.2-14B & 73.86 & 7 & $< 0.001$ & *** \\
\ourmodel vs. LTX-Video-ICLora & 70.26 & 7 & $< 0.001$ & *** \\
\ourmodel vs. UniAnimate-DiT & 67.45 & 7 & $< 0.001$ & *** \\
\ourmodel vs. Wan-2.2-Animate-14B & 40.89 & 7 & $< 0.001$ & *** \\
\bottomrule
\end{tabular}}
\end{table}

\section{More Visualizations of Photoreal Reference}
\label{sec:more_viz}

We provide additional video results generated from \ourmodel in our supplementary project page.

\noindent \textbf{Comparison to Previous Works on NeRSemble}
We compared the generation of \ourmodel to LTX-Video-0.9.8~\cite{HaCohen2024LTXVideo}, LTX-Video-IClora~\cite{HaCohen2024LTXVideo}, Wan2.2~\cite{wan2025}, UniAnimate-DiT~\cite{wang2025unianimate}, and Wan2.2-Animate~\cite{wan2025}. Noted that the text prompt input to LTX-Video-0.9.8 and Wan2.2 are captioned by Qwen-2.5-VL~\cite{bai2025qwen2} on the ground truth.

\noindent  \textbf{Photorealistic Video Animation}
We provide visualizations of more photorealistic hair motion generation. The reference input is generated by FLUX~\cite{flux2024}.

\clearpage
\newpage

\bibliographystyle{splncs04}
\bibliography{main}